\documentclass[compsoc]{IEEEtran}
\IEEEoverridecommandlockouts
\usepackage{cite}
\usepackage{caption}
\usepackage{amsmath,amssymb,amsfonts}
\usepackage{algorithm}
\usepackage[noend]{algpseudocode}
\usepackage{textcomp}
\usepackage{booktabs}
\usepackage[table,xcdraw]{xcolor}
\usepackage{comment}
\usepackage{array}
\def\BibTeX{{\rm B\kern-.05em{\sc i\kern-.025em b}\kern-.08em
    T\kern-.1667em\lower.7ex\hbox{E}\kern-.125emX}}
\usepackage{stfloats}
\usepackage{float}
\usepackage{hhline}
\usepackage[export]{adjustbox}

\usepackage[caption=false,font=footnotesize]{subfig}
\usepackage{graphicx}
\usepackage{multirow}
\usepackage{tabularx}

\usepackage{pgfplots}
\pgfplotsset{width=7cm,compat=1.8}
\begin{document}
\title{Towards Equitable ASD Diagnostics: A Comparative Study of Machine and Deep Learning Models Using Behavioral and Facial Data}

\author{Mohammed Aledhari, Mohamed Rahouti, and Ali Alfatemi
\thanks{M. Aledhari is with the Department of Information Science, University of North Texas, Denton, Texas, 76203 USA (e-mail: mohammed.aledhari@unt.edu).}

\thanks{M. Rahouti and A. Alfatemi are with the Department of Computer and Information Science, Fordham University, New York, NY, 10023 USA (e-mails: mrahouti@fordham.edu; aalfatemi@fordham.edu).}}

\IEEEtitleabstractindextext{%
    
    \begin{abstract}  

Autism Spectrum Disorder (ASD) is often underdiagnosed in females due to gender-specific symptom differences overlooked by conventional diagnostics. This study evaluates machine learning models, particularly Random Forest and convolutional neural networks, for enhancing ASD diagnosis through structured data and facial image analysis. Random Forest achieved 100\% validation accuracy across datasets, highlighting its ability to manage complex relationships and reduce false negatives, which is crucial for early intervention and addressing gender biases. In image-based analysis, MobileNet outperformed the baseline CNN, achieving 87\% accuracy, though a 30\% validation loss suggests possible overfitting, requiring further optimization for robustness in clinical settings. Future work will emphasize hyperparameter tuning, regularization, and transfer learning. Integrating behavioral data with facial analysis could improve diagnosis for underdiagnosed groups. These findings suggest Random Forest's high accuracy and balanced precision-recall metrics could enhance clinical workflows. MobileNet's lightweight structure also shows promise for resource-limited environments, enabling accessible ASD screening. Addressing model explainability and clinician trust will be vital.

    \end{abstract}
}

\maketitle
\IEEEdisplaynontitleabstractindextext
\section{Introduction}
Autism Spectrum Disorder (ASD) is a neurodevelopmental condition characterized by difficulties in social communication, interaction, and behavior, typically manifesting in early childhood \cite{maniscalco2020preliminary}. ASD affects approximately 1 in 36 children in the U.S., with boys diagnosed nearly 3.8 times more often than girls, leading to a gender-based diagnostic gap that frequently leaves females undiagnosed or misdiagnosed until later in life. Delays in detection can negatively impact long-term outcomes in communication, education, and mental health, contributing to significant economic costs; the annual financial burden of ASD in the U.S. was estimated at \$268 billion in 2015, projected to reach \$461 billion by 2025.

\begin{figure*}
    \centering
    \includegraphics[width=1\linewidth]{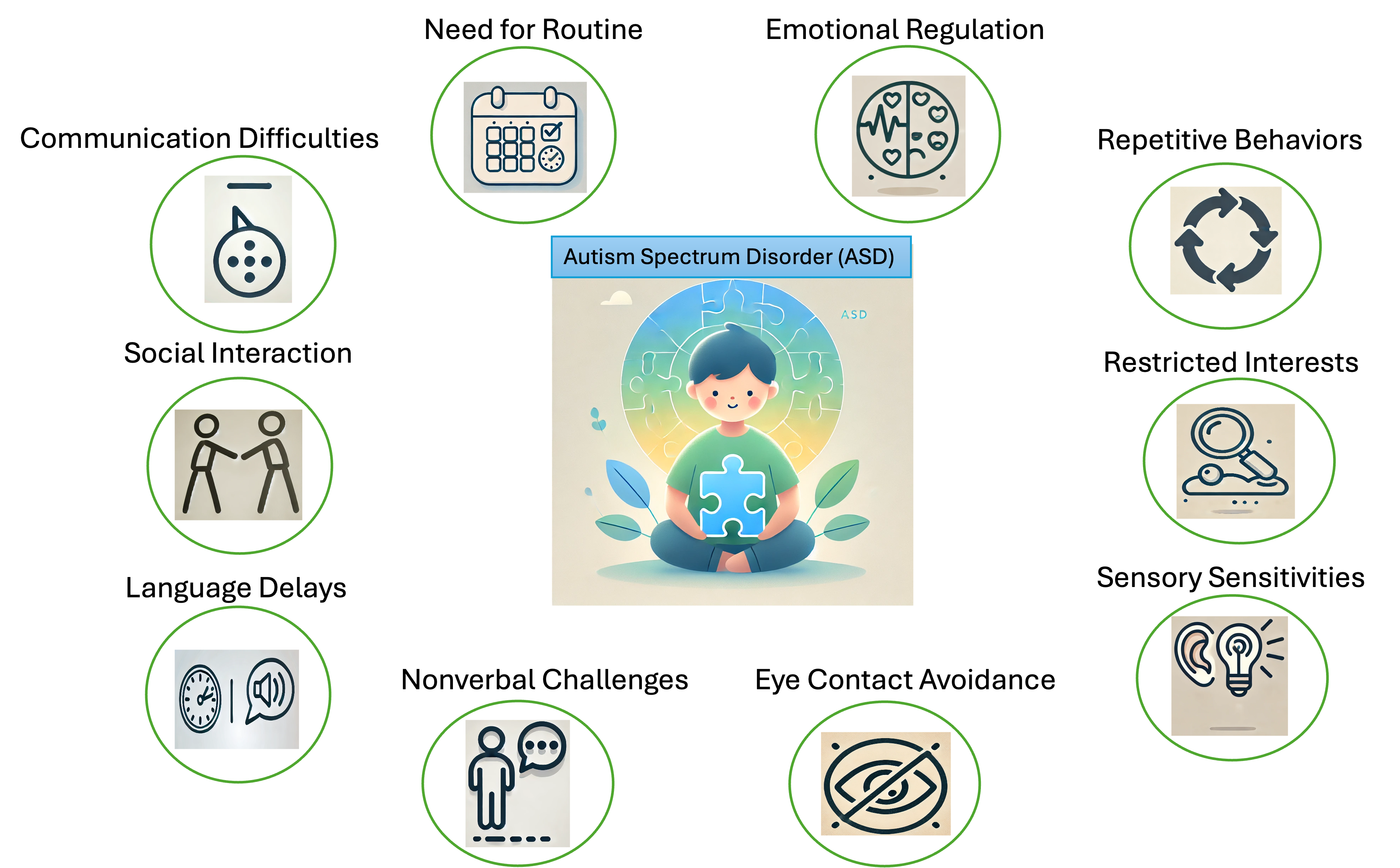}
    \caption{Visual representation of the key characteristics associated with ASD, highlighting the core aspects such as social interaction difficulties, repetitive behaviors, restricted interests, language delays, eye contact avoidance, communication difficulties, sensory sensitivities, nonverbal challenges, emotional regulation, and need for routine. The icons around the central figure illustrate the common behavioral and sensory traits often associated with ASD.}
    \label{fig-D}
\end{figure*}

These societal and personal impacts underscore the importance of recognizing and understanding the core symptoms of ASD, which often vary in their presentation across individuals and can complicate timely diagnosis. Fig.~\ref{fig-D} visually represents these key characteristics, including social interaction difficulties, repetitive behaviors, restricted interests, language delays, eye contact avoidance, communication difficulties, sensory sensitivities, nonverbal challenges, emotional regulation issues, and a need for routine.

The underdiagnosis of ASD in females can be attributed to several factors. Diagnostic criteria for ASD, which have been largely derived from studies focusing on male populations, may not adequately capture the more subtle or socially masked manifestations of the condition seen in females. Many females engage in "camouflaging," a strategy involving the suppression or masking of autistic traits in social settings, making clinical detection more challenging. Furthermore, existing diagnostic tools may not be sufficiently sensitive to gender-specific presentations of ASD, resulting in significant diagnostic delays. On average, females are diagnosed with ASD later in life than males, with some remaining undiagnosed well into adulthood.

Several theoretical models have attempted to explain these gender differences in autism presentation. The Female Autism Phenotype (FAP) theory suggests that females exhibit different behavioral characteristics compared to males, which may influence their diagnosis \cite{volkmar2021seeing}. Similarly, the Female Protective Effect (FPE) posits that females require a higher genetic or environmental burden to display comparable autistic traits as males \cite{zhang2020genetic}. Another influential theory, the Extreme Male Brain (EMB) hypothesis, posits that ASD may exaggerate cognitive patterns typically associated with males, driven by higher androgen levels \cite{xiong2020amniotic}. However, recent research has questioned the link between androgen levels and autism, highlighting the complexity of these theoretical frameworks.

Given these challenges, there is an urgent need for diagnostic tools that are more sensitive to the unique presentation of ASD in females. While traditional clinical methods are helpful, they may not fully capture the complexity of the condition, particularly in underdiagnosed populations. Advances in machine learning (ML) offer a promising solution. ML techniques can analyze large, complex datasets to detect patterns that may elude human evaluators, making them particularly suitable for identifying subtle differences in ASD presentations across genders. Several studies have demonstrated the utility of ML in diagnosing ASD and differentiating it from other neurodevelopmental conditions, such as attention deficit hyperactivity disorder (ADHD). However, the potential for ML to specifically improve diagnosis in females has been underexplored.

This study aims to address this gap by leveraging ML models to enhance the accuracy of ASD diagnosis in females. By applying convolutional neural networks (CNNs) to facial image data and traditional classifiers to behavioral data, we aim to identify gender-specific patterns that could improve early detection in females. Face detection, a subfield of computer vision, holds particular promise for this task, as facial features may differ between males and females with ASD, offering new biological insights. Although previous studies in this area have focused primarily on male subjects, this study seeks to extend these techniques to female populations, providing a novel approach to reducing the diagnostic gender gap.

In summary, current clinical guidelines for ASD diagnosis, mainly based on male presentations, are inconsistently applied, which disproportionately impacts females. ML, mainly when applied to facial and behavioral data, holds the potential to provide a scalable, objective, and sensitive method for diagnosing ASD in females. By evaluating the performance of various ML models across multiple datasets, this study seeks to improve early diagnosis and address the long-standing gender disparity in ASD detection.
\subsection{Research Problem}

ASD affects individuals across all genders, yet early diagnosis is essential for ensuring timely access to intervention, support, and tailored treatments. While early diagnosis can significantly improve long-term outcomes in communication, education, and mental health, current diagnostic tools are predominantly designed based on how ASD manifests in boys. This creates a significant gap in diagnosing females, whose symptoms may differ in presentation, often resulting in delayed diagnosis or misdiagnosis \cite{brunissen2021sex}. These diagnostic discrepancies prevent women from receiving appropriate early interventions, which are crucial for improving quality of life and functional outcomes.

Moreover, ASD diagnostic procedures are time-consuming and complex, often requiring multidisciplinary assessments and lengthy evaluations, which further delay access to care. Although ML has emerged as a potential tool to streamline diagnosis, relying solely on ML models is insufficient to capture the full range of ASD presentations across genders. A more integrated approach—combining ML with complementary techniques—is needed to ensure that gender-specific diagnostic nuances are identified and addressed effectively.

\subsection{Purpose of Study}

The purpose of this study is twofold: (1) to evaluate the limitations of current ASD diagnostic tools with a particular focus on their inability to detect ASD in females, and (2) to explore how face detection techniques can be applied to improve the accuracy and timeliness of ASD diagnosis across genders. We aim to investigate which facial features are most relevant for distinguishing ASD in males and females, thereby providing new insights into the biological and genetic factors that may influence ASD presentation. Specifically, this study seeks to answer the following research question: Can combining facial feature analysis with ML models enhance the accuracy of ASD diagnosis in females compared to traditional diagnostic methods?

\subsection{Motivation}

Accurate and timely diagnosis of ASD is crucial, as failing to diagnose ASD when it is present (a false-negative) can result in missed opportunities for early intervention, which is known to improve long-term developmental outcomes significantly. Additionally, a delayed diagnosis can leave individuals and families without necessary support and access to services. On the other hand, incorrectly diagnosing ASD (a false-positive) can lead to unnecessary treatments and diagnostic procedures, increasing emotional and financial strain on families and placing additional burdens on healthcare systems. The lack of a definitive medical test for ASD further complicates the diagnostic process, requiring clinicians to rely on observed behaviors and developmental history, which may not accurately reflect the experiences of all individuals—particularly females.

Many autistic women do not receive timely or accurate diagnoses, which hinders their ability to advocate for their needs and limits access to critical resources. Research has shown that receiving an ASD diagnosis later in life can significantly improve self-identity, mental health, and access to necessary accommodations. Our goal is to develop a more efficient and accurate method for identifying women who may be on the autism spectrum, enabling earlier intervention and empowering women to self-identify with greater confidence, even without a formal assessment.

\subsection{Contributions}

This study contributes to the growing field of ASD diagnosis by proposing an innovative approach that combines ML with face detection techniques to improve diagnostic accuracy, particularly for females. While face detection has been explored in ASD research, few studies have examined its potential to capture gender-specific diagnostic features. Our approach integrates facial analysis with ML models to provide a more nuanced understanding of ASD in males and females. By comparing a baseline CNN model with a more advanced CNN architecture, we explore the trade-offs between model complexity and diagnostic performance, offering insights into which models are most effective for this task. Our contributions include:
\begin{itemize}
    \item Developing a hybrid approach using facial and behavioral data for ASD diagnosis.
    \item Providing a comparative analysis of different CNN models for ASD classification.
    \item Highlighting the importance of biological and genetic factors in understanding ASD manifestation across genders.
\end{itemize}

Form the audience perspective, this research is designed for ML specialists, biologists, geneticists, and clinical practitioners interested in the intersection of technology and ASD diagnosis. For ML experts, the study provides a case study of how advanced models like CNNs can be applied to diagnostic challenges. Biologists and geneticists may find the focus on facial feature analysis and its potential links to ASD informative for their research on the biological basis of neurodevelopmental disorders. Clinicians and medical professionals will benefit from insights into how ML can support gender-sensitive diagnostic practices, particularly for underdiagnosed populations such as women.

\subsection{Paper Organization}

The remainder of this paper is organized as follows: Section 2 reviews the related literature on ASD diagnosis and the use of ML in healthcare. Section 3 presents our proposed method, combining ML and face detection techniques. Section 4 describes the experimental setup, including datasets and model configurations. Section 5 discusses the results of our experiments, and Section 6 concludes the paper with final thoughts and suggestions for future research.

\section{Background and Related Work}

Recent research has focused on leveraging whole-genome sequencing (WGS) combined with ML to predict antimicrobial resistance (AMR) effectively. Studies highlight how WGS data, when integrated with ML techniques such as random forests and convolutional neural networks, can accurately predict resistance patterns without relying solely on known resistance genes, a significant advancement for understudied pathogens. For example, a study on Escherichia coli demonstrated that encoding techniques like label encoding and frequency-based encoding enabled ML models to identify AMR by capturing mutations related to resistance beyond established gene markers \cite{Stoesser2013Predicting}. Another study found that WGS coupled with ML methods effectively predicted antimicrobial minimum inhibitory concentrations, especially for pathogens like Salmonella, achieving substantial accuracy in clinical isolate tests \cite{Porse2020Acinetobacter}. Further systematic reviews affirm that combining WGS with ML can streamline and enhance AMR prediction for critical pathogens, providing an adaptable framework for various bacterial species \cite{Wang2023Random}.

While whole-genome sequencing combined with ML has proven instrumental in identifying patterns of antimicrobial resistance, these techniques are promising in other areas of medical diagnostics, such as ASD. ML’s capacity to detect subtle patterns offers a similar advantage in ASD diagnostics, particularly in identifying gender-specific symptom manifestations often overlooked in traditional assessments. Research on the diagnosis of ASD in women has increased in recent years, with growing attention to how symptoms manifest differently between males and females. In parallel, advancements in ML and face detection offer promising avenues for improving the accuracy of ASD diagnosis, particularly in underdiagnosed populations such as women. This section reviews vital studies addressing gender differences in ASD presentation and explores how ML techniques have been applied to ASD diagnosis.

Incorporating advanced methodologies in ASD diagnosis is crucial for addressing the challenges posed by traditional approaches. Recent developments in ML for ASD diagnosis have explored innovative techniques. Albahri et al. \cite{albahri2023towards} present a novel model leveraging a T-spherical fuzzy-weighted method, offering a significant step toward enhancing diagnostic accuracy. Complementing this, Rasul et al. \cite{rasul2024evaluation} provide an evaluation of ML approaches for early ASD diagnosis. In contrast, Gao et al. \cite{gao2024comprehensive} explore deep learning models in multi-modal imaging markers for interpretation. Rasul et al. \cite{rasul2023early} further reinforce the importance of early ML diagnosis, and Alqaysi et al. \cite{alqaysi2024evaluation} introduce hybrid models utilizing a fuzzy sets-based decision-making model. 

Moreover, Bahathiq et al. \cite{bahathiq2024efficient} emphasize the efficacy of structural MRI in optimizing ML models. Parlett et al. \cite{parlett2023applications} review unsupervised ML applications, highlighting their potential in ASD research, while Cao and Cao \cite{cao2023commentary} reflect on the broader challenges and opportunities in this domain. Thapa et al. \cite{thapa2023machine} differentiate autism sub-classifications using ML, and Washington and Wall \cite{washington2023review} provide a roadmap for ASD-related data science research. Lastly, Alkahtani et al. \cite{alkahtani2023deep} focus on facial landmark-based deep learning for ASD identification, and Alves et al. \cite{alves2023diagnosis} explore functional brain networks combined with ML for improved diagnosis. These state-of-the-art approaches collectively advance the field of ASD diagnosis by leveraging ML's capability to detect nuanced patterns, providing a robust foundation for future research.

\subsection{Gender Differences in Behavior and Symptoms}

Several studies have investigated the differences in how ASD presents in males versus females. Lai et al. \cite{lai2015sex} proposed a four-level conceptual framework to explore the nosological and diagnostic challenges related to gender differences in ASD. This framework includes critical questions about how autism is defined, how gender influences the liability for developing ASD, and how developmental mechanisms differ between the sexes. Their work emphasizes the longstanding male bias in ASD research, which has led to a male-centered understanding of the disorder. While their framework provides a valuable tool for future studies, the complexity of the questions raised—such as the role of gender in autism development—has made it challenging to generate clear, unified findings.

Ratto et al. \cite{ratto2018girls} expanded on these ideas by using an IQ-matched sample of school-aged youth to assess sex differences in ASD traits and adaptive skills. Their results revealed that females with higher IQs were less likely to meet the criteria on the ADI-R diagnostic tool despite exhibiting more severe traits based on parent reports. Additionally, females outperformed males in language and vocabulary skills, suggesting they may be more adept at masking their symptoms. This raises concerns that current diagnostic tools, designed with male traits in mind, may be insufficiently sensitive to the subtler manifestations of ASD in females. The study highlights the need for diagnostic tools that account for these gender-based differences to reduce the high rate of false negatives in females.

Barbaro et al. \cite{barbaro2021investigating} focused on early signs of ASD in infants and toddlers, investigating gender differences in communication behaviors between 18 and 24 months. While they found no significant differences in this age range, the large gender ratio discrepancy (44 females to 153 males) suggests that females may require more careful monitoring to ensure early detection of ASD symptoms. The small sample size for females limits the generalizability of the findings, and future studies should address this limitation by increasing the female sample size. Nonetheless, their study underscores the importance of early detection, particularly in female infants who may otherwise be overlooked.

\subsection{Gender Biases in Diagnostic Tools}
The work by Randall et al. \cite{randall2018diagnostic} reviewed the effectiveness of several widely used ASD diagnostic tools, including ADOS, ADI-R, CARS, DISCO, GARS, and 3di, in diagnosing preschool children. Their findings showed that ADOS had the highest sensitivity, but all tools exhibited varying degrees of sensitivity and specificity, with none being fully reliable as a standalone assessment for ASD. The review also pointed out that these tools were developed with a male-biased understanding of ASD, potentially contributing to the underdiagnosis of females. The authors suggest that more gender-sensitive diagnostic criteria are needed, especially for preschool children, to ensure accurate and early diagnosis in both boys and girls.

Brown et al. \cite{brown2020autistic} proposed a modified version of the Girls Questionnaire for Autism Spectrum Condition (GQ-ASC) to address these biases. Their study, which included both cisgender and transgender women, identified five key components—imagination and play, camouflaging, sensory sensitivities, socializing, and interests—that were more relevant for diagnosing autism in women. By focusing on traits more common in females, the modified GQ-ASC achieved higher accuracy in distinguishing between autistic and non-autistic women. This work highlights the need for diagnostic tools that are sensitive to female-specific traits and demonstrates the importance of incorporating gender-related factors in ASD assessments.

\subsection{ML and Face Detection in ASD Diagnosis}
ML has proven to be a powerful tool in the medical domain, and recent research has applied it to the diagnosis of ASD, especially for analyzing complex behavioral and biological data. Studies using CNNs have shown that ML models can effectively analyze facial data to detect patterns associated with ASD. However, much of this research has focused on male-dominated datasets, limiting our understanding of how ASD manifests in females. Face detection, commonly used in computer vision tasks, has become a complementary tool in ASD diagnosis. By analyzing facial landmarks and structures, ML models can detect biological markers of ASD that may not be apparent in behavioral assessments. For example, studies have used CNNs to analyze facial features such as eye gaze, facial symmetry, and expression patterns, revealing significant differences between individuals with ASD and neurotypical controls. Despite these advancements, few studies have examined how face detection can be applied to address the diagnostic challenges specific to females with ASD, leaving a significant research gap.

The literature reveals several critical gaps in the diagnosis of ASD, particularly in females. While there is growing recognition of gender differences in ASD presentation, most diagnostic tools remain biased toward male characteristics, leading to underdiagnosis in women. Additionally, ML models, particularly those used in face detection, have shown promise in improving ASD diagnosis but have primarily focused on male populations. Our study builds on these findings by combining ML models and face detection techniques to explore ASD diagnosis in both males and females, specifically focusing on addressing the diagnostic gap in women. By integrating both facial and behavioral data, we aim to develop a more accurate and gender-sensitive diagnostic tool for ASD.
\subsection{Usage of ML and Face Detection in Diagnosing ASD}

Recent advances in ML and face detection have opened new possibilities for improving the diagnosis of ASD, mainly through analyzing facial morphology and applying computational models. This section reviews key studies that explore the relationship between facial features and ASD, as well as how ML techniques can enhance diagnostic accuracy, particularly in underrepresented groups such as females.

\subsection{Early Work on Facial Features and ASD Diagnosis}
Early research on facial features in ASD primarily focused on males. Tan et al. \cite{tan2017hypermasculinised} investigated whether boys and girls with ASD exhibit increased facial masculinity compared to typically developing controls. Using 3D photogrammetry, they analyzed facial features in a sample of 48 boys and 53 girls. Their study found that both boys and girls with ASD exhibited more androgynous facial features compared to neurotypical controls, suggesting a potential link between facial morphology and ASD traits. This study laid the foundation for subsequent research into how facial features might serve as diagnostic markers for ASD.

Building on these findings, Tan et al. \cite{tan2020broad} used a larger sample of 209 children to investigate facial masculinity in ASD further. This study used 3D facial photogrammetry to assess facial features in children, including non-autistic siblings of children with ASD. Their results indicated that both male and female siblings of children with ASD displayed more masculine facial features than neurotypical controls, reinforcing the idea that facial traits may have a genetic link to ASD. However, this study was limited to a Caucasian population, highlighting the need for further research across different ethnic groups to improve the generalizability of these findings.

\subsection{Gender Differences in Facial Morphology and ASD}
While much of the early work focused on male participants, more recent studies have explored gender-specific facial traits in ASD. Gilani et al. \cite{gilani2015sexually} identified six key facial features that distinguished males from females and then examined these features in individuals with varying levels of ASD traits. Using 3D facial images of 208 young adults, the study found that individuals with high levels of ASD traits tended to exhibit less sexually dimorphic facial features. This research suggests that facial morphology might reflect autistic traits differently in males and females, providing a potential tool for improving gender-sensitive ASD diagnoses.

Tan et al. \cite{tan2020sex} continued this line of inquiry by analyzing facial masculinity and femininity in neurotypical adults with varying levels of autistic traits. Their findings showed that higher levels of autistic traits were associated with decreased femininity in females and less pronounced masculinity in males. Although this study offered valuable insights into the relationship between facial morphology and autistic traits, it was limited to neurotypical individuals and did not include those with clinical ASD diagnoses. This highlights a significant gap in the literature, as studies focused on facial features in women with diagnosed ASD remain underexplored. Additionally, these studies have yet to consider other important variables like age or ethnicity, which may influence the generalizability of findings across populations.

\subsection{ML for ASD Diagnosis Using Facial Features}
The application of ML to ASD diagnosis has demonstrated promising results in recent years. One notable study by Jahanara et al. \cite{jahanara2021detecting} used CNNs to analyze facial features for ASD diagnosis. Their study applied transfer learning, using the VGG-19 architecture as the base model, and achieved an accuracy of 96\%. However, their model also exhibited a high validation loss of nearly 50\%, suggesting potential overfitting or issues with model generalization. While the accuracy results were impressive, the large gap between training and validation performance indicates that broader application requires further optimization.

Our study builds on this work by comparing the performance of a baseline CNN with a more efficient CNN architecture, MobileNet, which is designed to handle high-dimensional data with fewer parameters. By incorporating both facial and behavioral data, we aim to improve overall accuracy and reduce validation loss, making the model more generalizable. Additionally, our research addresses a key limitation of previous work by including both male and female participants, thereby reducing the gender bias prevalent in earlier studies. Another approach to ASD diagnosis using ML was presented by Alcañiz et al. \cite{alcaniz2020machine}, who combined ML with virtual reality to classify children with ASD based on their movement patterns. Using a virtual reality system, they tracked children’s body movements during various tasks and applied ML algorithms to classify ASD versus neurotypical children. While their study achieved an accuracy of 89\%, it suffered from a limited sample size, particularly among female participants. This reinforces the need for more inclusive research that accurately represents gender diversity in ASD diagnosis.

\subsection{Challenges and Opportunities in Using ML for ASD Diagnosis}
Despite the progress in applying ML and facial feature analysis to ASD diagnosis, several challenges remain. Many studies reviewed were conducted on male-dominated datasets, limiting the models' ability to generalize to female populations. Furthermore, facial feature analysis alone may not be sufficient for comprehensive ASD diagnosis, as behavioral and cognitive traits also play a critical role. Combining facial and behavioral data in ML models offers an opportunity to develop more accurate and gender-sensitive diagnostic tools. Our study aims to address these gaps by integrating facial feature analysis with behavioral data and applying advanced CNN models to both male and female participants. By doing so, we hope to improve the accuracy of ASD diagnosis across genders and contribute to a more nuanced understanding of how autistic traits are expressed in different populations.

\subsection{Uniqueness of this Study}

The existing body of work an ASD diagnosis has largely focused on male populations, overlooking gender-specific traits in females that contribute to underdiagnosis and misdiagnosis. Studies primarily utilize traditional diagnostic tools or ML models without adapting them to capture the subtle, often camouflaged symptoms more commonly seen in females with ASD. While some research has applied ML and facial feature analysis, these approaches typically lack a comprehensive representation of female subjects and fail to address gender biases in ASD diagnostic tools. This study uniquely bridges this gap by employing a hybrid approach, integrating both behavioral and facial data, emphasizing ML models tailored to identify gender-specific diagnostic patterns. By including a comparative analysis of Random Forest and MobileNet architectures, our study presents a more inclusive and nuanced framework that targets gender disparity in ASD diagnosis, enhancing early detection capabilities for underserved female populations.

\section{Proposed Method}

ASD diagnosis presents unique challenges, particularly in identifying subtle patterns in facial features that are often linked to ASD traits. Traditional diagnostic tools rely heavily on behavioral assessments, but recent advances in ML provide the potential to enhance these tools through automated image-based methods. This section proposes a method that applies CNNs to classify ASD from facial images. Our approach combines a standard CNN as a baseline with MobileNet, a more advanced and efficient architecture designed to optimize accuracy while minimizing computational costs.

\subsection{Baseline Method and Architecture Overview}

Our baseline model is a standard CNN, implemented via Keras, commonly used in image classification tasks. This baseline serves as a performance benchmark, allowing us to evaluate the effectiveness of more advanced architectures like MobileNet. The simplicity of the baseline CNN offers an initial understanding of how well a standard model can perform in classifying ASD from facial images. The typical CNN architecture is shown in Fig. \ref{fig:baselinecnn}.

\begin{figure}[h]
    \centering
    \includegraphics[width=0.5\textwidth]{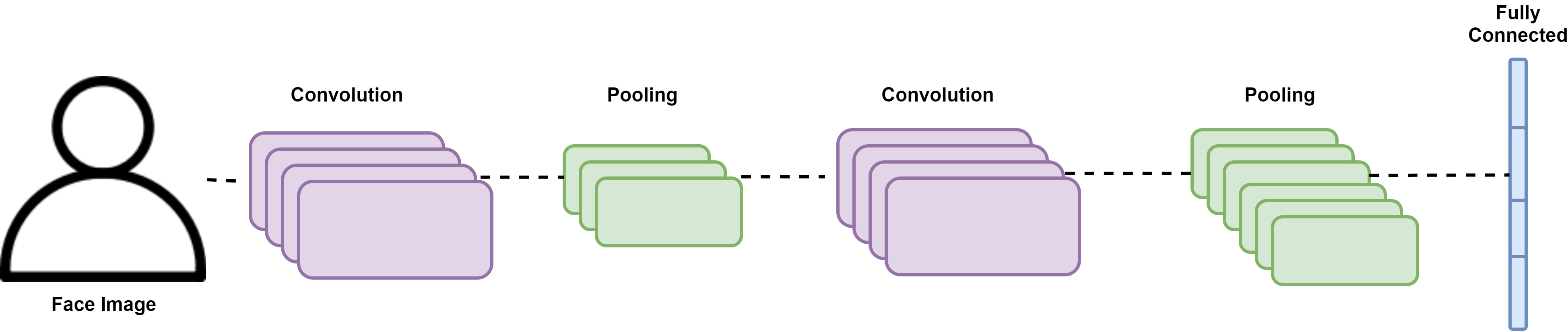}
    \caption{Typical CNN architecture.}
    \label{fig:baseline-cnn}
\end{figure}


\begin{itemize}
    \item The model consists of two convolutional layers, each followed by a ReLU activation function.
    \item A 2x2 max pooling layer reduces the spatial dimensions of the feature maps.
    \item A flattening layer converts the 2D feature maps into a 1D feature vector.
    \item A fully connected layer outputs the final classification, predicting whether the individual in the image has ASD or is neurotypical.
    \item Input images are resized to 64x64 pixels with three color channels (RGB), a standard input size for image classification tasks.
\end{itemize}

While the baseline CNN effectively extracts key features from facial images, its simple structure may limit its ability to capture more complex patterns that distinguish individuals with ASD. This motivates the need for a more sophisticated model, MobileNet, to perform better.

\subsection{Proposed Model: MobileNet}

Given the limitations of the baseline CNN, particularly in terms of computational efficiency and generalization, we propose using MobileNet, an architecture optimized for low-latency, resource-constrained environments. MobileNet, proposed by \cite{howard2017mobilenets}, is known for maintaining high accuracy with significantly fewer parameters than traditional CNNs, making it ideal for real-world applications such as ASD diagnosis in clinical or low-resource settings. The proposed modified MobileNet architecture is highlighted in Algorithm \ref{alg:MobileNet}.


\begin{algorithm}
\footnotesize
\caption{MobileNet architecture for ASD diagnosis.}
\label{alg:MobileNet}
\begin{algorithmic}[1]
\State \textbf{Input}: Training dataset $\mathcal{D} = \{(x_i, y_i)\}$ with facial images $x_i$ and labels $y_i$
\State \textbf{Objective}: Train MobileNet model $M_{\theta}$ to classify ASD from facial images with optimized accuracy and minimal overfitting

\Procedure{Data Preprocessing}{}
    \For{each image $x_i$ in $\mathcal{D}$}
        \State Resize $x_i$ to $224 \times 224 \times 3$
        \State Normalize pixel values of $x_i$ to range [0,1]
        \State Apply data augmentation (e.g., random rotation, flip, zoom)
    \EndFor
\EndProcedure

\Procedure{Initialize MobileNet Model}{}
    \State Define MobileNet architecture $M_{\theta}$ with depthwise separable convolutions
    \State Initialize weights $\theta$ using pre-trained MobileNet parameters
    \State Set model-specific hyperparameters:
    \State \hspace{\algorithmicindent} Batch size = 32
    \State \hspace{\algorithmicindent} Learning rate = 0.001 (with adaptive decay)
    \State \hspace{\algorithmicindent} Dropout rate = 0.5 for regularization
\EndProcedure

\Procedure{Training and Optimization}{}
    \For{each epoch $t = 1, \dots, T$}
        \State \textit{// Forward pass on training batch}
        \State Compute predictions $\hat{y}_i = M_{\theta}(x_i)$ for each $x_i$ in batch
        \State Calculate training loss $\mathcal{L}_{train} = \frac{1}{|B|} \sum_{x_i \in B} \text{CrossEntropy}(y_i, \hat{y}_i)$
        
        \State \textit{// Backward pass and parameter update}
        \State Update weights $\theta$ to minimize $\mathcal{L}_{train}$ using Adam optimizer
        
        \If{Validation loss has plateaued}
            \State Reduce learning rate by factor of 0.1
            \State Apply early stopping if no improvement for consecutive epochs
        \EndIf
    \EndFor
\EndProcedure

\Procedure{Evaluation}{}
    \State \textbf{Input}: Test dataset $\mathcal{D}_{test} = \{(x_j, y_j)\}$
    \For{each image $x_j$ in $\mathcal{D}_{test}$}
        \State Obtain prediction $\hat{y}_j = M_{\theta}(x_j)$
    \EndFor
    \State Compute evaluation metrics (accuracy, precision, recall, F1-score) on $\mathcal{D}_{test}$
\EndProcedure

\State \textbf{return} Trained MobileNet model $M_{\theta}$ with optimized parameters and evaluation metrics
\end{algorithmic}
\end{algorithm}

\subsection{Key Features of MobileNet}
\begin{itemize}
    \item \textbf{Depthwise separable convolutions:} Unlike standard convolutions, MobileNet replaces the conventional convolutional layers with depthwise separable convolutions, which split the spatial and depth dimensions. This results in a significant reduction in the number of parameters and computational complexity. For instance, a standard convolutional layer with 10 input channels, 20 output channels, and a 7x7 filter would require 2800 parameters. By separating the convolutional layer into depthwise and pointwise convolutions, MobileNet reduces the number of parameters to approximately 400, while still capturing essential spatial information.
    \item \textbf{Factorization for efficiency:} MobileNet’s architecture factorizes standard convolutions into smaller, more efficient operations. This reduces the risk of overfitting and enables the model to run efficiently on devices with limited processing power, such as smartphones or low-end clinical diagnostic systems. This feature is particularly relevant for deploying ASD diagnostic tools in resource-constrained environments.
\end{itemize}

\textbf{Why MobileNet?} The choice of MobileNet is driven by its ability to balance accuracy and efficiency, both of which are critical for real-time applications in clinical settings. While deeper models like DenseNet could capture more intricate patterns in large datasets, MobileNet is better suited for our use case due to its:
\begin{itemize}
    \item \textbf{Low latency:} MobileNet’s lightweight architecture allows it to process images quickly, which is essential for real-time diagnostic applications.
    \item \textbf{Scalability and deployment:} MobileNet is designed to work on various platforms, including low-power devices. This makes it ideal for ASD diagnosis tools that could be deployed in resource-limited environments, such as rural clinics or on mobile devices.
\end{itemize}

\subsection{Model Training and Preprocessing}

To train both the baseline CNN and MobileNet, we used a dataset of facial images of individuals with ASD and neurotypical individuals. Each image was preprocessed through a series of standard steps:
\begin{itemize}
    \item \textbf{Image normalization:} All images were normalized to a scale of [0, 1] by dividing pixel values by 255.
    \item \textbf{Data augmentation:} To reduce overfitting and improve generalization, data augmentation techniques such as random rotation, horizontal flipping, and zoom were applied. This ensured that the models were exposed to various image variations during training.
    \item \textbf{Batch size and learning rate:} We used a batch size of 32 and an adaptive learning rate schedule to fine-tune model performance and ensure convergence.
\end{itemize}

\subsection{Performance Expectations}

We expect MobileNet to outperform the baseline CNN in terms of both accuracy and efficiency. Specifically:
\begin{itemize}
    \item \textbf{Improved generalization:} Due to MobileNet’s use of depthwise separable convolutions, the model is expected to generalize better to unseen data, particularly in capturing subtle facial features linked to ASD without overfitting.
    \item \textbf{Reduced overfitting:} By reducing the number of parameters, MobileNet is less likely to overfit on the training data compared to the baseline CNN.
    \item \textbf{Faster inference:} MobileNet’s lightweight architecture will likely result in faster inference times, making it suitable for real-time ASD diagnosis applications.
\end{itemize}

The proposed method leverages the strengths of two CNN architectures: a standard baseline CNN for benchmarking and MobileNet for enhanced performance and efficiency. By incorporating depthwise separable convolutions and optimizing for low-latency environments, MobileNet addresses the limitations of the baseline model, offering a more practical solution for ASD diagnosis in clinical and real-world applications. The following sections will provide an in-depth analysis of both models' experimental setups, results, and performance metrics.



\section{Experimental Setup}

The experimental setup of this study was designed to evaluate the performance of both the baseline CNN and the MobileNet architecture for diagnosing ASD from facial images and structured datasets. The following sections outline the implementation, research questions, dataset selection, preprocessing techniques, hyperparameter tuning, and evaluation metrics.

\subsection{Implementation Details}

Our models were implemented in Python 3.7 using TensorFlow and Keras, widely recognized for their flexibility in deep learning tasks. We chose these libraries due to their rich support for building, training, and evaluating neural network architectures. JetBrains PyCharm was utilized as the Integrated Development Environment (IDE) due to its convenient debugging and project management capabilities.

To speed up the training process, we leveraged CUDA (Compute Unified Device Architecture) for GPU acceleration, running the experiments on an ASUS Laptop with an NVIDIA RTX 2060 GPU. The GPU significantly reduced the training time by allowing for efficient parallel computation, particularly during the more complex MobileNet architecture training. CUDA enabled simultaneous processing of larger image batches, improving model convergence. The performance gain was critical in achieving rapid iterations during hyperparameter tuning.

\subsection{Research Questions}

The following research questions were developed to guide the experimental design, focusing on the potential of photographic analysis to contribute to autism diagnosis. These questions helped shape the selection of datasets and evaluation metrics, ensuring a comprehensive and meaningful analysis.
\subsubsection{Can photographic analysis of facial images reliably diagnose autism?}
This question evaluates the feasibility and effectiveness of using image-based models, such as CNNs, to identify ASD patterns in facial images. The goal is to compare the results from the image-based diagnosis with those from traditional structured data-based models.
\subsubsection{Can photographic analysis of facial images help address the underdiagnosis of autism in females?}
Since females are often underdiagnosed due to less pronounced symptoms, this question investigates whether the facial image analysis approach can improve detection rates in females, who are typically overlooked by conventional diagnostic methods.
\subsubsection{What specific facial features (if any) should be considered when analyzing facial images for diagnosing autism?}
This question explores the potential for identifying specific facial features that correlate with ASD diagnosis, enabling the refinement of diagnostic models and criteria for both males and females.

These questions guided the selection of datasets and evaluation metrics, ensuring our analysis is comprehensive and meaningful.

\subsection{Dataset and Preprocessing Techniques}

\begin{table}[h]
\centering
\caption{Datasets used for model training, testing, and validation}
\label{tab:datasets}
\begin{tabularx}{\linewidth}{|l|X|c|}
\hline
\textbf{Dataset Type}         & \textbf{Details}                                                                                             & \textbf{Records/Images} \\ \hline
\multirow{3}{*}{Structured Data} & Children dataset (ages 4–11)                                                                              & 292                      \\ \cline{2-3} 
                              & Adolescent dataset (ages 12–16)                                                                           & 104                      \\ \cline{2-3} 
                              & Adult dataset (ages 18+)                                                                                  & 704                      \\ \hline
\multirow{3}{*}{Facial Image Data}  & Kaggle dataset - Training set                                                                          & 2,536                    \\ \cline{2-3} 
                              & Kaggle dataset - Testing set                                                                             & 200                      \\ \cline{2-3} 
                              & Kaggle dataset - Validation set                                                                          & 100                      \\ \hline
\end{tabularx}
\end{table}

As summarized in Table \ref{tab:datasets}, we used four datasets for training, testing, and validating our models. These datasets include both structured data and facial images, ensuring a comprehensive evaluation across multiple modalities.

\subsubsection{Structured datasets}
Three datasets were sourced from the UCI ML repository and published by \cite{thabtah2017autism}, covering different age groups-children, adolescents, and adults-to ensure the model's ability to generalize across a broad demographic. The children dataset includes 292 records (ages 4–11), the adolescent dataset comprises 104 records (ages 12–16), and the adult dataset contains 704 records (ages 18+). Each dataset includes 21 attributes relevant to ASD diagnosis, such as social communication patterns and behavioral traits, supporting model performance across developmental stages.
\subsubsection{Facial image dataset}
The Kaggle facial image dataset contains a total of 2,833 images labeled as either autistic or neurotypical. For model development, these images were divided into three subsets: a training set with 2,536 images, a testing set with 200 images, and a validation set with 100 images.

The images have a resolution of 224x224 pixels with three color channels (RGB). The dataset was preprocessed to remove duplicates and organized into subdirectories for training, testing, and validation, each containing autistic and non-autistic images.

\subsection{Preprocessing Techniques:}
\begin{itemize}
    \item \textbf{Image resizing:} All images were resized to 224x224x3 to meet the input size requirements of MobileNet.
    \item \textbf{Normalization:} Pixel values were normalized to the [0,1] range by dividing by 255.
    \item \textbf{Data augmentation:} To prevent overfitting and improve the model's ability to generalize, data augmentation techniques, such as random rotations, horizontal flipping, and zoom, were applied.
    \item \textbf{Handling class imbalance:} Although the Kaggle dataset was balanced, we reviewed the structured datasets for class imbalance. We applied techniques like class weighting during model training to address potential imbalances and used additional data augmentation for the minority class when necessary.
\end{itemize}

\subsection{Hyperparameter Tuning}

Both the baseline CNN and MobileNet models were tuned using the following hyperparameters:
\begin{itemize}
    \item \textbf{Optimizer:} Adam optimizer was chosen for both models due to its adaptive learning rate and efficient convergence.
    \item \textbf{Learning rate:} An initial learning rate of 0.001 was used, with an adaptive decay schedule to fine-tune the model as training progressed.
    \item \textbf{Batch size:} A batch size of 32 was selected to balance memory usage and computational efficiency.
    \item \textbf{Number of epochs:} The models were trained for 50 epochs, with early stopping applied based on validation accuracy to prevent overfitting.
    \item \textbf{Regularization:} L2 regularization was applied to the fully connected layers to prevent overfitting. Additionally, dropout (with a rate of 0.5) was applied to mitigate overfitting during training.
\end{itemize}

Hyperparameters were optimized through grid search and manual tuning, ensuring the best performance across datasets.

\begin{table}[h]
\centering
\caption{ML model results with 70\% training and 30\% testing.}
\label{tab:table1}
\begin{tabular}{@{}cc@{}}
\toprule
\multicolumn{2}{c}{\cellcolor[HTML]{DAE8FC}\textbf{ML Models Results (70\% Training 30\% Testing)}} \\ \midrule
\multicolumn{2}{c}{\textbf{Validation Accuracy - Children with ASD}}                                              \\ \hline
Support Vector Machine                                           & 0.90                                           \\
Random Forest                                                    & 1.0                                            \\
Gaussian Naive Bayes                                             & 0.94                                           \\
KNN                                                              & 0.86                                           \\ \hline
\multicolumn{2}{c}{\textbf{Validation Accuracy - Adults with ASD}}                                                \\ \hline
Support Vector Machine                                           & 0.89                                           \\
Random Forest                                                    & 0.99                                           \\
Gaussian Naive Bayes                                             & 0.98                                           \\
KNN                                                              & 0.91                                           \\ \hline
\multicolumn{2}{c}{\textbf{Validation Accuracy - Adolescents with ASD}}                                           \\ \hline
Support Vector Machine                                           & 0.90                                           \\
Random Forest                                                    & 1.0                                            \\
Gaussian Naive Bayes                                             & 0.84                                           \\
KNN                                                              & 0.90                                           \\ \bottomrule
\end{tabular}
\end{table}

\begin{table}[h]
\centering
\caption{ML model results with 80\% training and 20\% testing.}
\label{tab:table2}
\begin{tabular}{@{}cc@{}}
\toprule
\multicolumn{2}{c}{\cellcolor[HTML]{DAE8FC}\textbf{ML Models Results (80\% Training 20\% Testing)}} \\ \midrule
\multicolumn{2}{c}{\textbf{Validation Accuracy - Children with ASD}}                                              \\ \hline
Support Vector Machine                                           & 0.91                                           \\
Random Forest                                                    & 1.0                                            \\
Gaussian Naive Bayes                                             & 0.98                                           \\
KNN                                                              & 0.91                                           \\ \hline
\multicolumn{2}{c}{\textbf{Validation Accuracy - Adults with ASD}}                                                \\ \hline
Support Vector Machine                                           & 0.89                                           \\
Random Forest                                                    & 1.0                                            \\
Gaussian Naive Bayes                                             & 0.97                                           \\
KNN                                                              & 0.92                                           \\ \hline
\multicolumn{2}{c}{\textbf{Validation Accuracy - Adolescents with ASD}}                                           \\ \hline
Support Vector Machine                                           & 0.90                                           \\
Random Forest                                                    & 1.0                                            \\
Gaussian Naive Bayes                                             & 0.90                                           \\
KNN                                                              & 0.90                                           \\ \bottomrule
\end{tabular}
\end{table}

\begin{table}[h]
\centering
\caption{ML model results with 90\% training and 10\% testing.}
\begin{tabular}{@{}cc@{}}
\toprule
\multicolumn{2}{c}{\cellcolor[HTML]{DAE8FC}\textbf{ML Models Results (90\% Training 10\% Testing)}} \\ \midrule
\multicolumn{2}{c}{\textbf{Validation Accuracy - Children with ASD}}                                              \\ \hline
Support Vector Machine                                           & 0.90                                           \\
Random Forest                                                    & 1.0                                            \\
Gaussian Naive Bayes                                             & 0.96                                           \\
KNN                                                              & 0.93                                           \\ \hline
\multicolumn{2}{c}{\textbf{Validation Accuracy - Adults with ASD}}                                                \\ \hline
Support Vector Machine                                           & 0.88                                           \\
Random Forest                                                    & 1.0                                            \\
Gaussian Naive Bayes                                             & 0.97                                           \\
KNN                                                              & 0.90                                           \\ \hline
\multicolumn{2}{c}{\textbf{Validation Accuracy - Adolescents with ASD}}                                           \\ \hline
Support Vector Machine                                           & 1.0                                            \\
Random Forest                                                    & 1.0                                            \\
Gaussian Naive Bayes                                             & 0.81                                           \\
KNN                                                              & 0.90                                           \\ \bottomrule
\end{tabular}
\end{table}

\subsection{Evaluation Metrics for Model's Performance}

To assess the model's performance comprehensively, we used the following metrics. These metrics were chosen to evaluate not only the model's accuracy but also its ability to generalize to new data and handle potential classification challenges specific to ASD diagnosis.
\subsubsection{Accuracy}
This metric represents the overall percentage of correct predictions, evaluating performance on both the training and validation sets.
\subsubsection{Validation Accuracy}
Validation accuracy assesses how well the model generalizes to unseen data by measuring performance on the validation set. A close match between training and validation accuracy is critical for determining whether the model is overfitting.
\subsubsection{Loss and Validation Loss}
Loss measures the model’s error during training, while validation loss tracks performance on the validation set. These metrics were essential for detecting overfitting or underfitting.
\subsubsection{Precision, Recall, and F1-Score}
While accuracy is important, the potential consequences of misclassification in ASD diagnosis require additional, detailed metrics to ensure reliability. Precision, recall, and the F1-score provide insight into the model's performance in correctly classifying ASD cases.

Precision measures the proportion of true positives out of all positive predictions, while recall measures the ability to identify true positives among all actual positives. The F1 score provides a balanced measure of precision and recall, particularly useful when the classes are imbalanced.

The experimental setup was carefully designed to evaluate the performance of CNN-based models in diagnosing ASD using both structured datasets and facial images. We ensured a comprehensive model performance analysis by focusing on key metrics such as accuracy, loss, precision, recall, and F1-score. Additionally, preprocessing techniques and hyperparameter tuning were applied to optimize the models and prevent overfitting. The following section will discuss the results and analysis of both the baseline CNN and MobileNet models.

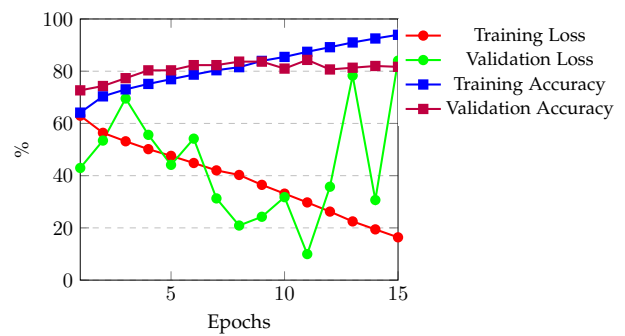
\begin{figure}[h!]
    \centering
        \resizebox{\columnwidth}{!}{ 
    \begin{tikzpicture}
        \begin{axis}[
            title={Training and Validation Loss and Accuracy (SGD Optimizer)},
            xlabel={Epochs},
            ylabel={\%},
            xmin=1, xmax=15,
            ymin=0, ymax=100,
            legend style={font=\small, at={(1,1)}, anchor=north west, draw=none},
            ymajorgrids=true,
            grid style=dashed
        ]
        
        \addplot[
            color=red,
            mark=*,
            line width=1pt
        ]
        coordinates {
            (1, 62.96) (2, 56.40) (3, 53.16) (4, 50.15) (5, 47.58)
            (6, 44.85) (7, 42.02) (8, 40.27) (9, 36.48) (10, 33.12)
            (11, 29.75) (12, 26.24) (13, 22.48) (14, 19.43) (15, 16.38)
        };
        \addlegendentry{Training Loss}
        
        \addplot[
            color=green,
            mark=*,
            line width=1pt
        ]
        coordinates {
            (1, 42.95) (2, 53.44) (3, 69.55) (4, 55.63) (5, 44.14)
            (6, 54.17) (7, 31.29) (8, 20.90) (9, 24.26) (10, 31.77)
            (11, 9.97) (12, 35.76) (13, 78.32) (14, 30.68) (15, 84.04)
        };
        \addlegendentry{Validation Loss}
        
        \addplot[
            color=blue,
            mark=square*,
            line width=1pt
        ]
        coordinates {
            (1, 64.14) (2, 70.37) (3, 73.04) (4, 75.13) (5, 76.94)
            (6, 78.69) (7, 80.37) (8, 81.54) (9, 83.95) (10, 85.47)
            (11, 87.43) (12, 89.19) (13, 91.01) (14, 92.53) (15, 93.93)
        };
        \addlegendentry{Training Accuracy}
        
        \addplot[
            color=purple,
            mark=square*,
            line width=1pt
        ]
        coordinates {
            (1, 72.67) (2, 74.33) (3, 77.33) (4, 80.33) (5, 80.33)
            (6, 82.33) (7, 82.33) (8, 83.67) (9, 83.67) (10, 81.00)
            (11, 84.33) (12, 80.67) (13, 81.33) (14, 82.00) (15, 81.67)
        };
        \addlegendentry{Validation Accuracy}
        
        \end{axis}
    \end{tikzpicture}
    }
    \caption{Baseline CNN results with SGD optimizer showing combined Training Loss, Validation Loss, Training Accuracy, and Validation Accuracy over 15 epochs.}
    \label{fig:latex1}
\end{figure}

\begin{figure}[h]
    \centering
    \includegraphics[width=0.5\textwidth]{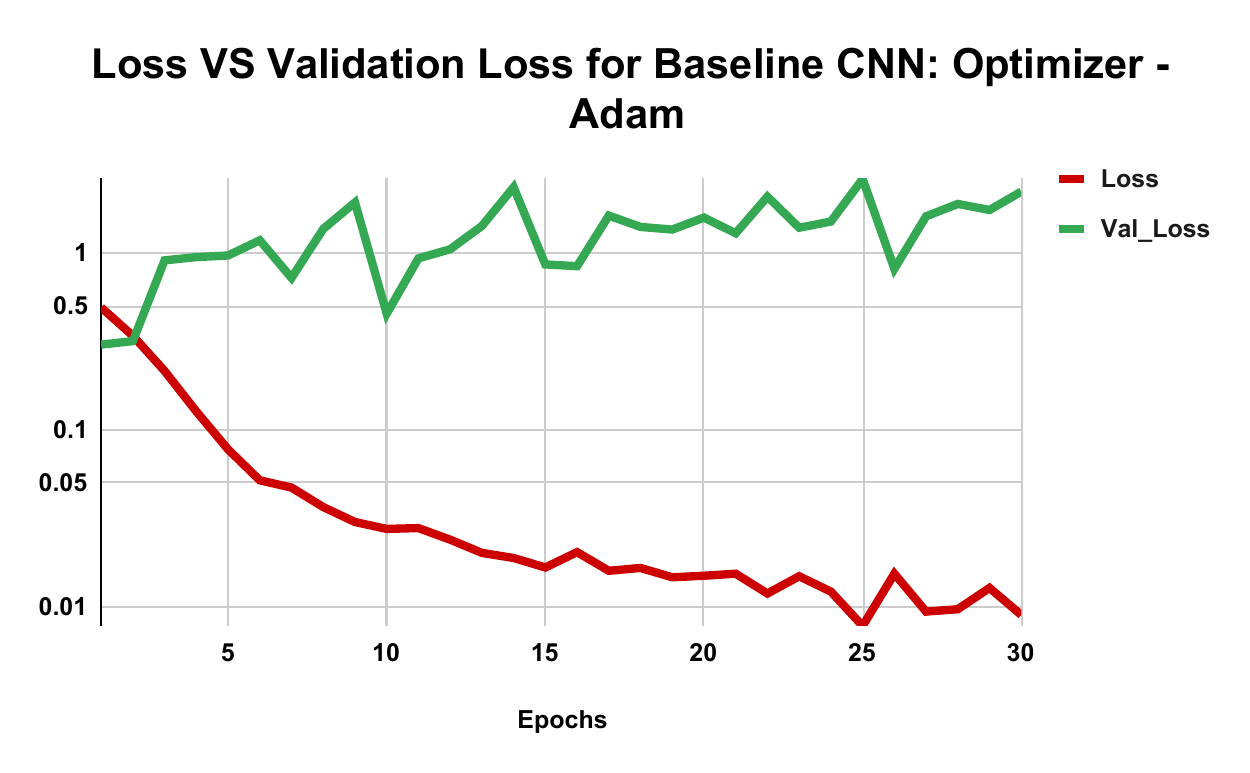}
    \caption{Loss results from the second test run for the baseline CNN with the Adam optimizer. The green line represents the validation loss.}
    \label{fig:baseline-cnn-loss-test-2-adam}
\end{figure}

\begin{figure}[h]
    \centering
    \includegraphics[width=0.5\textwidth]{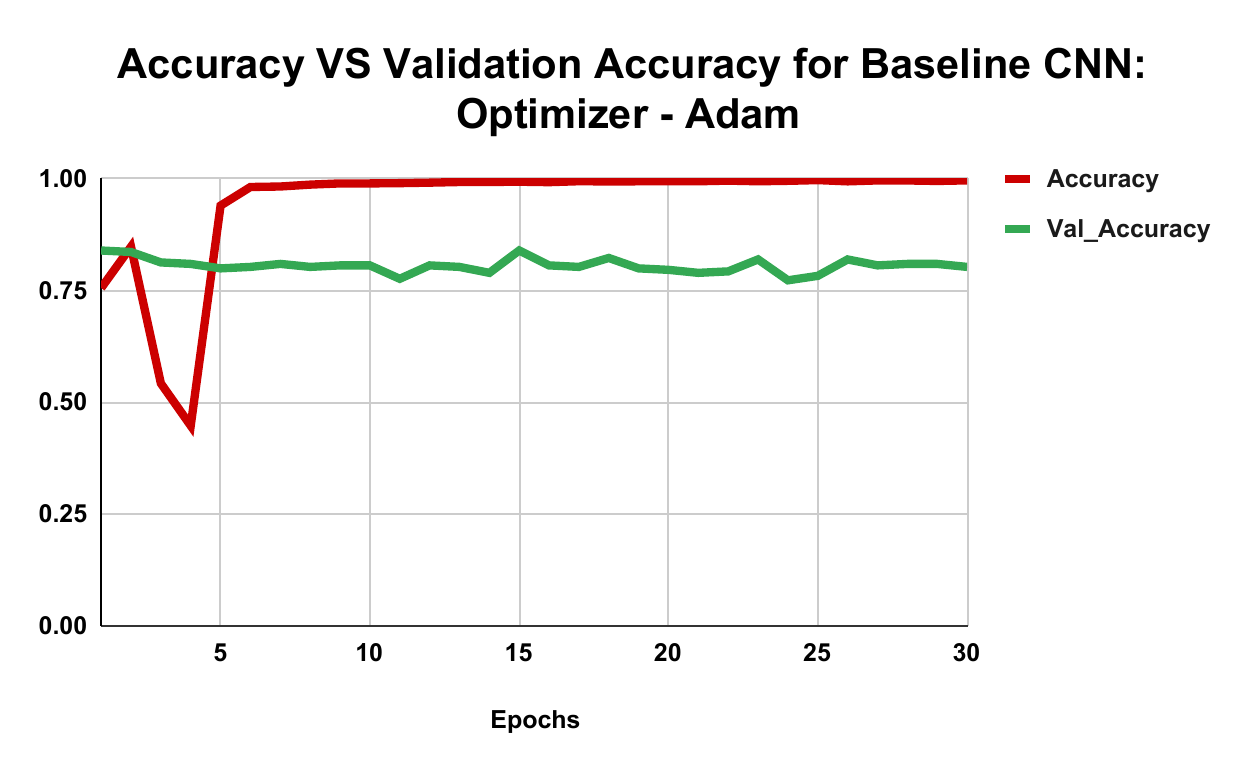}
    \caption{Accuracy results from the second test run for the baseline CNN with the Adam optimizer.}
    \label{fig:baseline-cnn-accuracy-test-2-adam}
\end{figure}


\section{Results and Discussion}
The results of this study, as shown in Table \ref{fig:latex1}, clearly indicate that Random Forest consistently outperformed all other models across various training-testing splits and datasets. It achieved 100\% validation accuracy in all configurations (Tables 1, 2, and 3), underscoring its robustness in handling non-linear relationships and complex high-dimensional data. This exceptional performance highlights Random Forest’s ability to generalize effectively, regardless of dataset size or complexity. Its ensemble learning approach aggregates multiple decision trees, allowing the model to capture intricate patterns in ASD-related data more effectively than simpler models. Random Forest is an ideal candidate for structured datasets in Autism research, where early and accurate diagnosis is crucial. The high precision and recall observed in Random Forest's performance minimize the chances of false positives and false negatives, which are critical factors in clinical settings, ensuring that both over-diagnosis and missed diagnoses are avoided.

On the other hand, Support Vector Machine (SVM), as demonstrated in Tables 1, 2, and 3, achieved respectable accuracy, ranging from 0.89 to 0.91, but it fell short of Random Forest’s perfect accuracy. SVM performed relatively well in high-dimensional spaces, a trait that aligns with its known strengths in binary classification tasks. However, its performance declined in more complex datasets, suggesting that SVM may struggle to capture the nuanced interdependencies in Autism-related data. Furthermore, the kernel selection process in SVM adds complexity, as the model’s effectiveness is highly dependent on selecting the right kernel (e.g., linear or radial basis function). Despite these limitations, SVM remains a useful baseline model for simpler tasks or less feature-dense datasets. However, its lower recall compared to Random Forest implies a higher likelihood of missed diagnoses, which can delay intervention—a critical factor in Autism research and treatment.

K-Nearest Neighbors (KNN) delivered competitive results, with validation accuracies ranging from 0.86 to 0.93 across different datasets (Tables 1, 2, and 3). KNN showed particular strength in the children’s and adolescent datasets, achieving 0.93 and 0.91 validation accuracy, respectively. This suggests that KNN can be an effective tool for datasets with fewer dimensions or simpler feature distributions. However, its performance diminished in more complex datasets due to the curse of dimensionality, a well-known issue where KNN struggles as the number of features increases. This is particularly problematic in ASD-related datasets, which are typically feature-rich and involve complex interactions between behavioral and biological markers. Although KNN’s simplicity and interpretability make it valuable for exploratory analyses, its limitations in handling large and complex datasets reduce its practical utility in real-world Autism diagnosis tasks, where high-dimensional data are common.

In contrast, Gaussian Naive Bayes demonstrated the weakest performance, particularly in the adolescent dataset, which only achieved 0.81 validation accuracy (Tables 1, 2, and 3). This decline is likely due to the model’s assumption of feature independence, which is not well-suited for the highly interdependent features found in Autism data. Despite its efficiency in smaller, simpler datasets, Gaussian Naive Bayes struggled to capture the intricate relationships between features in more complex datasets, leading to poorer overall performance. Its highest accuracy of 0.90 in the children’s dataset suggests that it may be useful for preliminary screening or in datasets with minimal feature interdependence. However, its limitations make it unsuitable for nuanced diagnostic tasks where capturing the relationships between features is essential for accurate classification.

\begin{table}[h]
\centering
\caption{Baseline CNN results with Adam optimizer.}
\label{tab:Adam optimizer}
\begin{tabular}{@{}ccccc@{}}
\toprule
\multicolumn{5}{c}{\cellcolor[HTML]{DAE8FC}\textbf{Baseline CNN Results: Optimizer - Adam}}       \\ \midrule
\textbf{Epochs} & \textbf{Loss} & \textbf{Val\_Loss} & \textbf{Accuracy} & \textbf{Val\_Accuracy} \\ \hline
1               & 0.5314        & 0.3644             & 0.7300            & 0.7800                 \\
2               & 0.4216        & 0.8300             & 0.8036            & 0.8200                 \\
3               & 0.3054        & 0.4511             & 0.8684            & 0.8267                 \\
4               & 0.1877        & 0.9615             & 0.9247            & 0.8133                 \\
5               & 0.1163        & 0.4932             & 0.9573            & 0.8300                 \\
6               & 0.0792        & 1.5086             & 0.9718            & 0.8100                 \\
7               & 0.0568        & 0.0556             & 0.9806            & 0.7933                 \\
8               & 0.0472        & 0.7978             & 0.9837            & 0.7767                 \\
9               & 0.0370        & 2.0558             & 0.9872            & 0.8300                 \\
10              & 0.0419        & 0.7362             & 0.9856            & 0.8067                 \\
11              & 0.0345        & 0.9047             & 0.9880            & 0.8000                 \\
12              & 0.0258        & 0.4358             & 0.9917            & 0.8033                 \\
13              & 0.0269        & 0.2676             & 0.9907            & 0.7700                 \\
14              & 0.0295        & 0.9084             & 0.9902            & 0.7967                 \\
15              & 0.0166        & 2.1164             & 0.9943            & 0.8133                 \\ \bottomrule
\end{tabular}
\end{table}

\begin{figure}[h]
    \centering
    \includegraphics[width=0.5\textwidth]{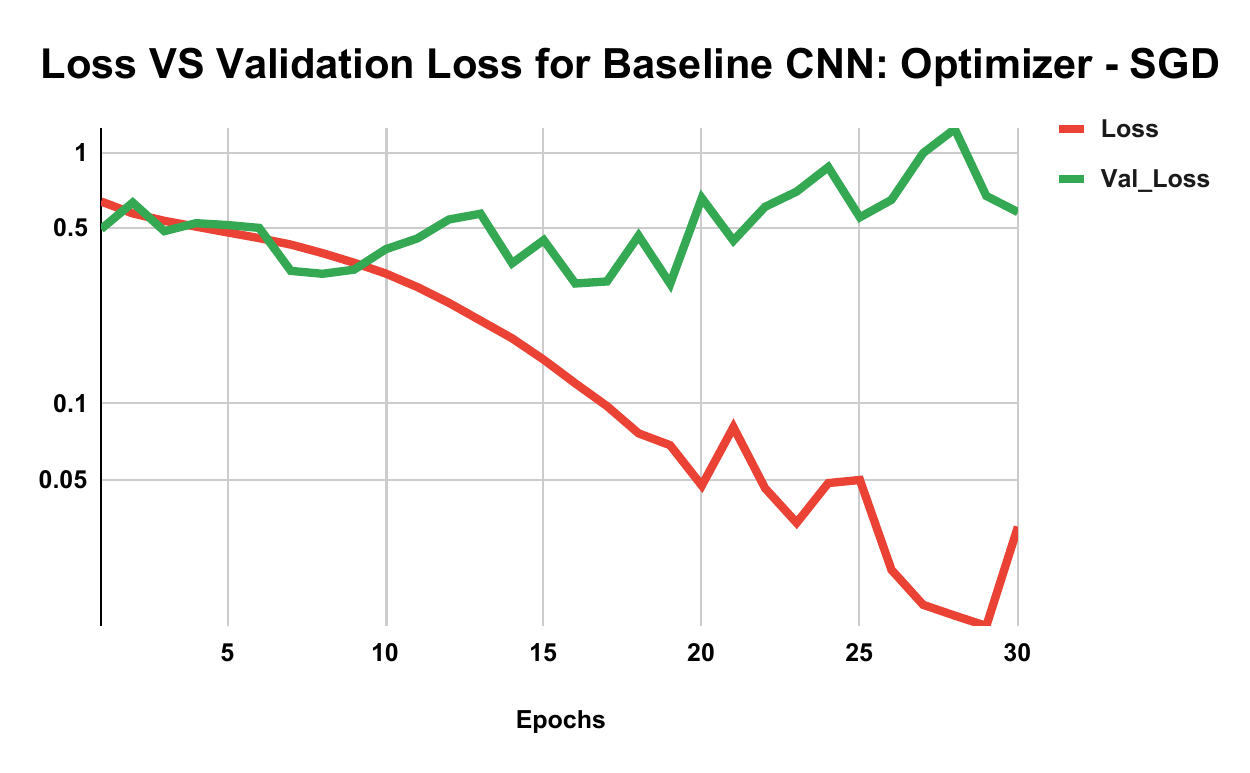}
    \caption{Loss results from the second test run for the baseline CNN with the SGD optimizer.}
    \label{fig:baseline-cnn-loss-test-2-sgd}
\end{figure}

\begin{figure}[h]
    \centering
    \includegraphics[width=0.5\textwidth]{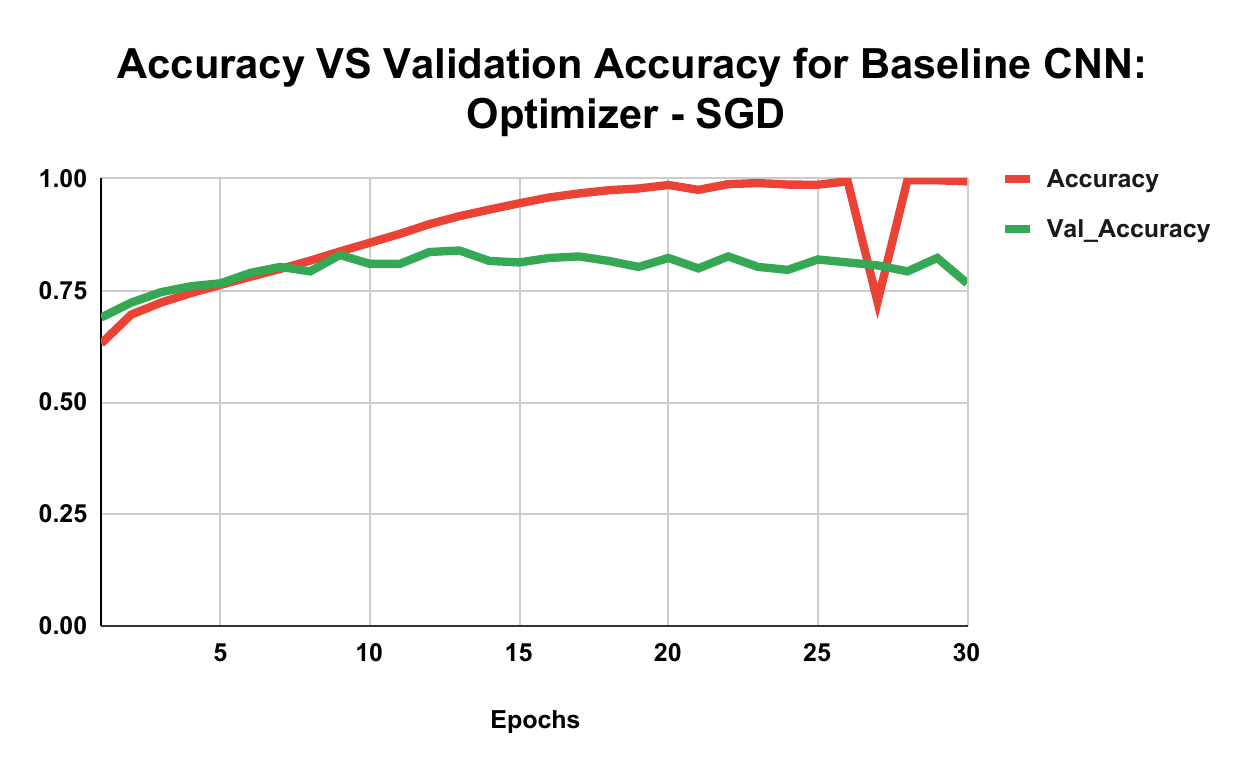}
    \caption{Accuracy results from the second test run for the baseline CNN with the SGD optimizer.}
    \label{fig:baseline-cnn-accuracy-test-2-sgd}
\end{figure}

\begin{figure}[h]
    \centering
    \includegraphics[width=0.45\textwidth]{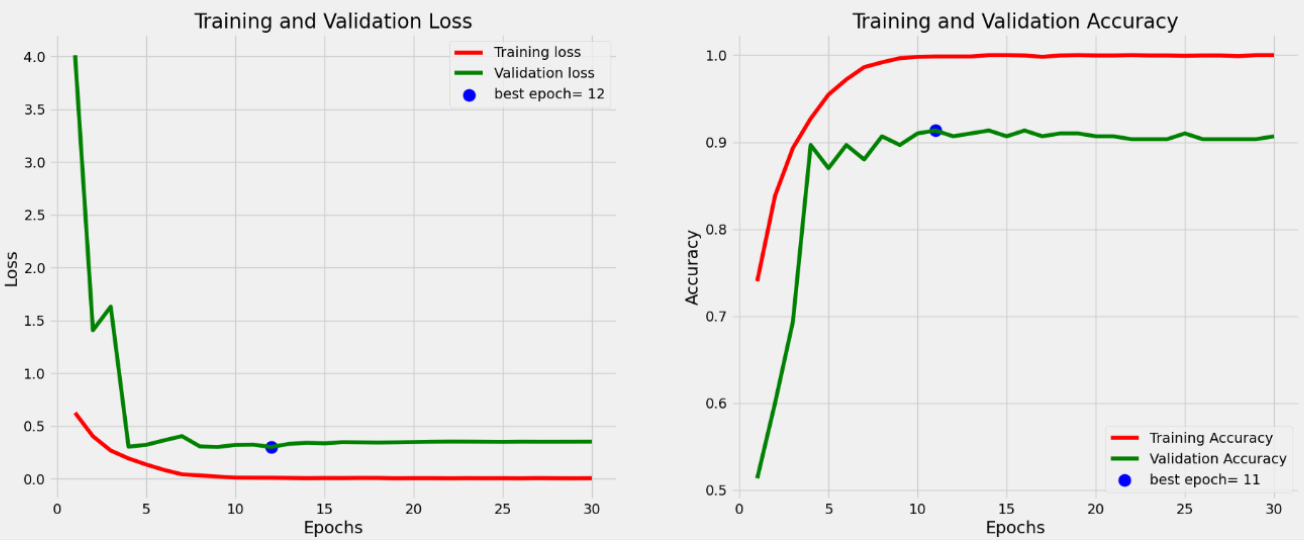}
    \caption{Results from the first CNN MobileNet model test run. The red line represents training, while the green line represents validation. The best-performing epoch is plotted in blue. Here, the best performing epochs were epoch 12 and epoch 11. Epoch 12 achieved the lowest validation loss of about 30\% while Epoch 11 achieved the highest validation accuracy of 91\%.}
    \label{fig:mobilenet-cnn1}
\end{figure}

\begin{figure}[h]
    \centering
    \includegraphics[width=0.45\textwidth]{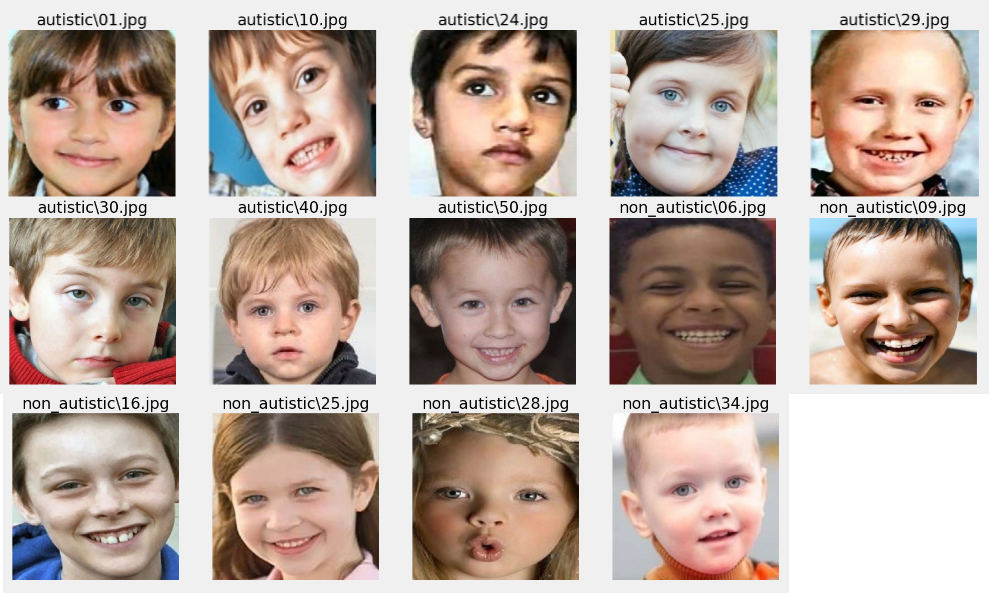}
    \caption{Photos that were miss-classified from the CNN MobileNet model on the first test run. There were 14 misclassified photos in total.}
    \label{fig:autism-missclassified-1}
\end{figure}

\begin{figure}[h!]
    \centering
    \includegraphics[width=0.45\textwidth]{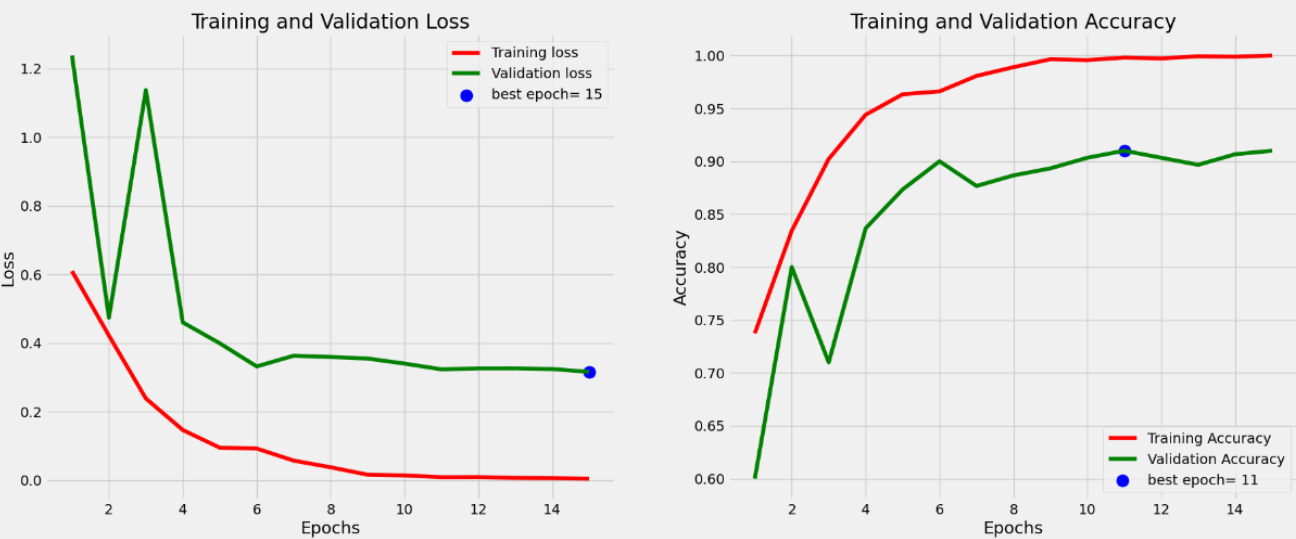}
    \caption{Results from the second CNN MobileNet model test run. In this second test run, the best-performing epochs were Epochs 15 and 11. Epoch 11 achieved a validation accuracy of 91\% and a validation loss of 32\%. Epoch 15 achieved a validation loss of 33\% and a validation accuracy of 90\%.}
    \label{fig:mobilenet-cnn2}
\end{figure}

\begin{figure}[h]
    \centering
    \includegraphics[width=0.45\textwidth]{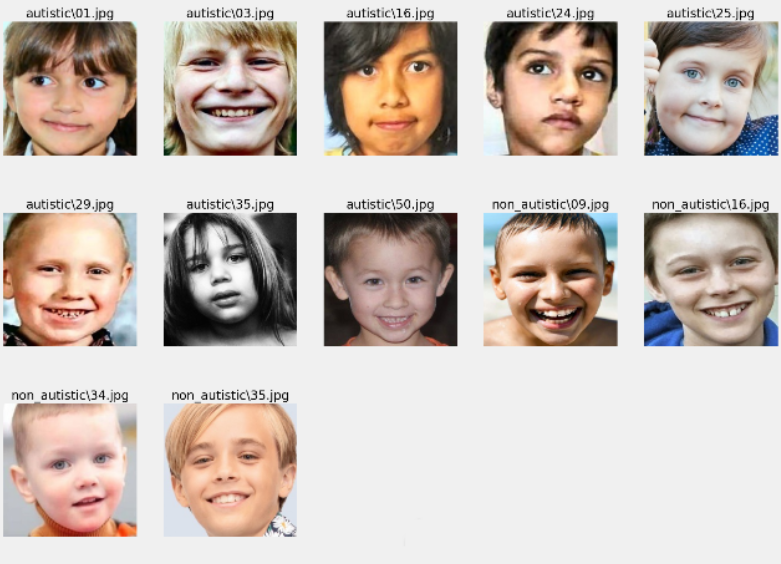}
    \caption{Photos that were miss-classified from the CNN MobileNet model on the second test run. There were 12 misclassified photos in total.}
    \label{fig:autism-missclassified-2}
\end{figure}

The results from the CNNs, particularly MobileNet, offer insights into the potential of deep learning models in Autism diagnosis. The baseline CNN's performance fluctuated between 77\% and 83\% validation accuracy when using the Adam optimizer as shown in Table \ref{tab:Adam optimizer}, with significant instability in validation loss, which ranged from 0.26 to 1.50 (Figures \ref{fig:baseline-cnn-loss-test-2-adam} and \ref{fig:baseline-cnn-accuracy-test-2-adam}). This suggests that the baseline CNN may be prone to overfitting and unable to generalize well to unseen data. MobileNet, a more advanced architecture, performed better, achieving 87\% validation accuracy (Fig. \ref{fig:mobilenet-cnn1}), but still exhibited a high validation loss of approximately 30\%, indicating potential overfitting as shown in Figures \ref{fig:autism-missclassified-1}, \ref{fig:mobilenet-cnn2}, and \ref{fig:autism-missclassified-2}. These results highlight the need for further tuning of the CNN models, particularly in applying more advanced regularization techniques like dropout or L2 regularization to improve generalization. Despite these challenges, MobileNet's ability to handle high-dimensional data efficiently makes it a promising candidate for real-world applications, especially in resource-constrained environments where diagnostic efficiency is crucial.

A key takeaway from the optimizer comparison is the trade-off between stability and performance. The Adam optimizer led to better initial accuracy as shown in Table \ref{tab:Adam optimizer}, but the model's instability in loss values indicated that it may not have been optimally tuned for this specific task (Figures \ref{fig:baseline-cnn-loss-test-2-adam} and \ref{fig:baseline-cnn-accuracy-test-2-adam}). In contrast, the SGD optimizer produced more stable results (Fig. \ref{fig:latex1}), with validation accuracy peaking at 84\% and a smoother loss trend (Figures \ref{fig:baseline-cnn-loss-test-2-sgd} and \ref{fig:baseline-cnn-accuracy-test-2-sgd}). This suggests that for tasks involving ASD diagnosis, training stability might be more critical than achieving rapid convergence and that further experimentation with learning rates and decay schedules could enhance performance with Adam or SGD.

In broader terms, the findings of this study have significant implications for clinical applications. Random Forest’s superior performance suggests it could be effectively integrated into clinical workflows for early and accurate Autism diagnosis, minimizing the likelihood of both false positives and false negatives. This is particularly valuable for underdiagnosed populations, such as females, where traditional diagnostic tools often fail to capture subtle behavioral patterns. Furthermore, MobileNet’s efficiency and high accuracy make it a viable option for use in mobile diagnostic tools or resource-constrained environments, such as rural clinics or areas with limited access to advanced medical facilities. However, its high validation loss underscores the need for further tuning before deployment in real-world clinical settings.

Future research should enhance these models through regularization, hyperparameter tuning, and transfer learning from pre-trained models such as ResNet or InceptionNet. This could significantly improve the performance of CNN-based models in Autism diagnosis. Additionally, exploring hybrid approaches that combine the strengths of Random Forest and CNN architectures could provide even more accurate and generalizable diagnostic tools, particularly for complex datasets involving structured and image-based data. In conclusion, the findings from this study lay a strong foundation for the use of ML models in Autism research while also pointing to critical areas for further improvement and application in clinical practice.

\section{Conclusion} 
This study confirms the transformative potential of ML models in diagnosing ASD using both structured datasets and facial image analysis. Random Forest consistently demonstrated 100\% accuracy across all datasets, underscoring its robustness and reliability in classifying ASD, even in complex, high-dimensional data environments. This exceptional performance is significant in clinical settings where minimizing diagnostic errors is critical. The high precision and recall of Random Forest make it especially suitable for ensuring that ASD diagnoses are accurate and timely, reducing both false negatives and false positives, which can significantly affect treatment outcomes. In the image-based analysis, MobileNet outperformed the baseline CNN, achieving a validation accuracy of 87\%, though its high validation loss (\~30\%) highlights the need for further optimization to enhance generalizability. Addressing this overfitting through hyperparameter tuning and regularization will ensure the model’s efficacy in real-world clinical settings, particularly in Autism diagnosis, where the stakes of misclassification are high. Despite these challenges, MobileNet’s ability to efficiently handle high-dimensional data shows promise, particularly for deployment in resource-constrained environments such as rural clinics or mobile health units, where quick and accurate ASD screening is essential. Future work must focus on several critical improvements.

The models must be validated on larger, more diverse datasets, such as the Kaggle facial image dataset, to ensure their robustness extends across different data types and populations. Furthermore, reducing MobileNet’s validation loss will improve its practical application in clinical settings. Advanced techniques like transfer learning from pre-trained models (e.g., ResNet or InceptionNet) could enhance model performance, allowing the system to leverage previously learned patterns and reduce computational costs. Recurrent Neural Networks (RNNs) should also be explored for facial image analysis, as RNNs could extract sequential information that may be relevant for detecting subtle changes in facial expressions, which are often overlooked in standard approaches. One of the most significant contributions of this research lies in addressing gender disparities in ASD diagnosis. Females are frequently underdiagnosed, and our approach of combining behavioral data with facial analysis aims to mitigate this issue by providing a more comprehensive diagnostic framework. This integration could enhance the accuracy of ASD detection in underdiagnosed groups, such as females, by capturing a broader spectrum of symptoms that traditional tools often miss. Expanding the dataset diversity and incorporating larger, more representative samples will ensure these models generalize across different demographics and populations, ultimately making them more inclusive and equitable. Once fully optimized, these ML models will have significant real-world applications. MobileNet’s lightweight architecture, for example, is particularly suited for resource-constrained environments, offering a practical solution for rapid and accurate ASD diagnosis in clinics that lack advanced equipment. 

Integrating these models into existing diagnostic workflows allows healthcare providers to access automated, data-driven insights that enhance early diagnosis, especially in underrepresented and underserved populations. This could dramatically improve outcomes by enabling timely intervention when it is most effective. Our long-term goal is to develop a robust and accessible diagnostic tool that integrates facial image analysis and behavioral data and can be deployed across various healthcare settings. By addressing current model limitations, such as overfitting in CNNs, and continuing to refine and test these models on diverse datasets, we hope to create a diagnostic system that is not only accurate but also scalable and accessible. Such tools could have far-reaching impacts on global healthcare, improving early ASD diagnosis, especially in communities with limited access to specialized care. Through these advancements, we aim to contribute to better, more personalized treatment strategies for individuals affected by ASD, ultimately improving their long-term outcomes.

\bibliographystyle{IEEEtran}
\bibliography{References}

\begin{thebibliography}{10}
\providecommand{\url}[1]{#1}
\csname url@samestyle\endcsname
\providecommand{\newblock}{\relax}
\providecommand{\bibinfo}[2]{#2}
\providecommand{\BIBentrySTDinterwordspacing}{\spaceskip=0pt\relax}
\providecommand{\BIBentryALTinterwordstretchfactor}{4}
\providecommand{\BIBentryALTinterwordspacing}{\spaceskip=\fontdimen2\font plus
\BIBentryALTinterwordstretchfactor\fontdimen3\font minus \fontdimen4\font\relax}
\providecommand{\BIBforeignlanguage}[2]{{%
\expandafter\ifx\csname l@#1\endcsname\relax
\typeout{** WARNING: IEEEtran.bst: No hyphenation pattern has been}%
\typeout{** loaded for the language `#1'. Using the pattern for}%
\typeout{** the default language instead.}%
\else
\language=\csname l@#1\endcsname
\fi
#2}}
\providecommand{\BIBdecl}{\relax}
\BIBdecl

\bibitem{maniscalco2020preliminary}
L.~Maniscalco, B.-B. Fr{\'e}d{\'e}rique, M.~Roccella, D.~Matranga, and G.~Tripi, ``A preliminary study on cranio-facial characteristics associated with minor neurological dysfunctions (mnds) in children with autism spectrum disorders (asd),'' \emph{Brain Sciences}, vol.~10, no.~8, p. 566, 2020.

\bibitem{volkmar2021seeing}
F.~R. Volkmar, M.~Woodbury-Smith, S.~L. Macari, and R.~A. {\O}ien, ``Seeing the forest and the trees: Disentangling autism phenotypes in the age of dsm-5,'' \emph{Development and Psychopathology}, pp. 1--9, 2021.

\bibitem{zhang2020genetic}
Y.~Zhang, N.~Li, C.~Li, Z.~Zhang, H.~Teng, Y.~Wang, T.~Zhao, L.~Shi, K.~Zhang, K.~Xia \emph{et~al.}, ``Genetic evidence of gender difference in autism spectrum disorder supports the female-protective effect,'' \emph{Translational psychiatry}, vol.~10, no.~1, pp. 1--10, 2020.

\bibitem{xiong2020amniotic}
H.~Xiong and S.~Scott, ``Amniotic testosterone and psychological sex differences: A systematic review of the extreme male brain theory,'' \emph{Developmental Review}, vol.~57, p. 100922, 2020.

\bibitem{brunissen2021sex}
L.~Brunissen, E.~Rapoport, K.~Chawarska, and A.~Adesman, ``Sex differences in gender-diverse expressions and identities among youth with autism spectrum disorder,'' \emph{Autism Research}, vol.~14, no.~1, pp. 143--155, 2021.

\bibitem{Stoesser2013Predicting}
N.~Stoesser, E.~M. Batty, D.~W. Eyre, M.~Morgan, D.~H. Wyllie, C.~Del Ojo~Elias \emph{et~al.}, ``Predicting antimicrobial susceptibilities for escherichia coli and klebsiella pneumoniae isolates using whole genomic sequence data,'' \emph{Journal of Antimicrobial Chemotherapy}, vol.~68, pp. 2234--2244, 2013.

\bibitem{Porse2020Acinetobacter}
A.~Porse, L.~J. Jahn, M.~M. Ellabaan, and M.~O. Sommer, ``Machine learning and feature extraction for rapid antimicrobial resistance prediction of acinetobacter baumannii from whole-genome sequencing data,'' \emph{Frontiers in Microbiology}, 2020.

\bibitem{Wang2023Random}
C.-C. Wang, Y.-T. Hung, C.-Y. Chou, S.-L. Hsuan, Z.-W. Chen, P.-Y. Chang, T.-R. Jan, and C.-W. Tung, ``Using random forest to predict antimicrobial minimum inhibitory concentrations of nontyphoidal salmonella in taiwan,'' \emph{Veterinary Research}, vol.~54, no.~1, p.~11, 2023.

\bibitem{albahri2023towards}
A.~S. Albahri, A.~A. Zaidan, H.~A. AlSattar, R.~A. Hamid, O.~S. Albahri, S.~Qahtan, and A.~H. Alamoodi, ``Towards physician's experience: Development of machine learning model for the diagnosis of autism spectrum disorders based on complex t-spherical fuzzy-weighted zero-inconsistency method,'' \emph{Computational Intelligence}, vol.~39, no.~2, pp. 225--257, 2023.

\bibitem{rasul2024evaluation}
R.~A. Rasul, P.~Saha, D.~Bala, S.~R.~U. Karim, M.~I. Abdullah, and B.~Saha, ``An evaluation of machine learning approaches for early diagnosis of autism spectrum disorder,'' \emph{Healthcare Analytics}, vol.~5, p. 100293, 2024.

\bibitem{gao2024comprehensive}
J.~Gao, Y.~Xu, Y.~Li, F.~Lu, and Z.~Wang, ``Comprehensive exploration of multi-modal and multi-branch imaging markers for autism diagnosis and interpretation: insights from an advanced deep learning model,'' \emph{Cerebral Cortex}, vol.~34, no.~2, p. bhad521, 2024.

\bibitem{rasul2023early}
R.~A. Rasul, P.~Saha, D.~Bala, S.~Karim, I.~Abdullah, and B.~Saha, ``Early diagnosis of autism spectrum disorder using machine learning approaches,'' \emph{arXiv preprint arXiv:2309.11646}, 2023.

\bibitem{alqaysi2024evaluation}
M.~Alqaysi, A.~Albahri, and R.~A. Hamid, ``Evaluation and benchmarking of hybrid machine learning models for autism spectrum disorder diagnosis using a 2-tuple linguistic neutrosophic fuzzy sets-based decision-making model,'' \emph{Neural Computing and Applications}, vol.~36, no.~29, pp. 18\,161--18\,200, 2024.

\bibitem{bahathiq2024efficient}
R.~A. Bahathiq, H.~Banjar, S.~K. Jarraya, A.~K. Bamaga, and R.~Almoallim, ``Efficient diagnosis of autism spectrum disorder using optimized machine learning models based on structural mri,'' \emph{Applied Sciences}, vol.~14, no.~2, p. 473, 2024.

\bibitem{parlett2023applications}
C.~M. Parlett-Pelleriti, E.~Stevens, D.~Dixon, and E.~J. Linstead, ``Applications of unsupervised machine learning in autism spectrum disorder research: a review,'' \emph{Review Journal of Autism and Developmental Disorders}, vol.~10, no.~3, pp. 406--421, 2023.

\bibitem{cao2023commentary}
X.~Cao and J.~Cao, ``Commentary: Machine learning for autism spectrum disorder diagnosis--challenges and opportunities--a commentary on schulte-r{\"u}ther et al.(2022),'' \emph{Journal of Child Psychology and Psychiatry}, vol.~64, no.~6, pp. 966--967, 2023.

\bibitem{thapa2023machine}
R.~Thapa, A.~Garikipati, M.~Ciobanu, N.~Singh, E.~Browning, J.~DeCurzio, G.~Barnes, F.~Dinenno, Q.~Mao, and R.~Das, ``Machine learning differentiation of autism spectrum sub-classifications,'' \emph{Journal of Autism and Developmental Disorders}, pp. 1--16, 2023.

\bibitem{washington2023review}
P.~Washington and D.~P. Wall, ``A review of and roadmap for data science and machine learning for the neuropsychiatric phenotype of autism,'' \emph{Annual review of biomedical data science}, vol.~6, no.~1, pp. 211--228, 2023.

\bibitem{alkahtani2023deep}
H.~Alkahtani, T.~H. Aldhyani, and M.~Y. Alzahrani, ``Deep learning algorithms to identify autism spectrum disorder in children-based facial landmarks,'' \emph{Applied Sciences}, vol.~13, no.~8, p. 4855, 2023.

\bibitem{alves2023diagnosis}
C.~L. Alves, T.~G. d.~O. Toutain, P.~de~Carvalho~Aguiar, A.~M. Pineda, K.~Roster, C.~Thielemann, J.~A.~M. Porto, and F.~A. Rodrigues, ``Diagnosis of autism spectrum disorder based on functional brain networks and machine learning,'' \emph{Scientific Reports}, vol.~13, no.~1, p. 8072, 2023.

\bibitem{lai2015sex}
M.-C. Lai, M.~V. Lombardo, B.~Auyeung, B.~Chakrabarti, and S.~Baron-Cohen, ``Sex/gender differences and autism: setting the scene for future research,'' \emph{Journal of the American Academy of Child \& Adolescent Psychiatry}, vol.~54, no.~1, pp. 11--24, 2015.

\bibitem{ratto2018girls}
A.~B. Ratto, L.~Kenworthy, B.~E. Yerys, J.~Bascom, A.~T. Wieckowski, S.~W. White, G.~L. Wallace, C.~Pugliese, R.~T. Schultz, T.~H. Ollendick \emph{et~al.}, ``What about the girls? sex-based differences in autistic traits and adaptive skills,'' \emph{Journal of autism and developmental disorders}, vol.~48, no.~5, pp. 1698--1711, 2018.

\bibitem{barbaro2021investigating}
J.~Barbaro and N.~C. Freeman, ``Investigating gender differences in the early markers of autism spectrum conditions (asc) in infants and toddlers,'' \emph{Research in Autism Spectrum Disorders}, vol.~83, p. 101745, 2021.

\bibitem{randall2018diagnostic}
M.~Randall, K.~J. Egberts, A.~Samtani, R.~J. Scholten, L.~Hooft, N.~Livingstone, K.~Sterling-Levis, S.~Woolfenden, and K.~Williams, ``Diagnostic tests for autism spectrum disorder (asd) in preschool children,'' \emph{Cochrane Database of Systematic Reviews}, no.~7, 2018.

\bibitem{brown2020autistic}
C.~M. Brown, T.~Attwood, M.~Garnett, and M.~A. Stokes, ``Am i autistic? utility of the girls questionnaire for autism spectrum condition as an autism assessment in adult women,'' \emph{Autism in Adulthood}, vol.~2, no.~3, pp. 216--226, 2020.

\bibitem{tan2017hypermasculinised}
D.~W. Tan, S.~Z. Gilani, M.~T. Maybery, A.~Mian, A.~Hunt, M.~Walters, and A.~J. Whitehouse, ``Hypermasculinised facial morphology in boys and girls with autism spectrum disorder and its association with symptomatology,'' \emph{Scientific reports}, vol.~7, no.~1, pp. 1--11, 2017.

\bibitem{tan2020broad}
D.~W. Tan, M.~T. Maybery, S.~Z. Gilani, G.~A. Alvares, A.~Mian, D.~Suter, and A.~J. Whitehouse, ``A broad autism phenotype expressed in facial morphology,'' \emph{Translational psychiatry}, vol.~10, no.~1, pp. 1--9, 2020.

\bibitem{gilani2015sexually}
S.~Z. Gilani, D.~W. Tan, S.~N. Russell-Smith, M.~T. Maybery, A.~Mian, P.~R. Eastwood, F.~Shafait, M.~Goonewardene, and A.~J. Whitehouse, ``Sexually dimorphic facial features vary according to level of autistic-like traits in the general population,'' \emph{Journal of neurodevelopmental disorders}, vol.~7, no.~1, pp. 1--10, 2015.

\bibitem{tan2020sex}
D.~W. Tan, M.~T. Maybery, L.~Ewing, J.-X. Tay, P.~R. Eastwood, and A.~J. Whitehouse, ``Sex-specific variation in facial masculinity/femininity associated with autistic traits in the general population,'' \emph{British Journal of Psychology}, vol. 111, no.~4, pp. 723--741, 2020.

\bibitem{jahanara2021detecting}
S.~Jahanara and S.~Padmanabhan, ``Detecting autism from facial image,'' 2021.

\bibitem{alcaniz2020machine}
M.~Alca{\~n}iz~Raya, J.~Mar{\'\i}n-Morales, M.~E. Minissi, G.~Teruel~Garcia, L.~Abad, and I.~A. Chicchi~Giglioli, ``Machine learning and virtual reality on body movements’ behaviors to classify children with autism spectrum disorder,'' \emph{Journal of clinical medicine}, vol.~9, no.~5, p. 1260, 2020.

\bibitem{howard2017mobilenets}
A.~G. Howard, M.~Zhu, B.~Chen, D.~Kalenichenko, W.~Wang, T.~Weyand, M.~Andreetto, and H.~Adam, ``Mobilenets: Efficient convolutional neural networks for mobile vision applications,'' 2017.

\bibitem{thabtah2017autism}
F.~Thabtah, ``Autism spectrum disorder screening: machine learning adaptation and dsm-5 fulfillment,'' in \emph{Proceedings of the 1st International Conference on Medical and health Informatics 2017}, 2017, pp. 1--6.

\end{thebibliography}

\end{document}